%% file: main.tex
\documentclass[letterpaper, 10 pt, conference]{ieeeconf}

\IEEEoverridecommandlockouts
\usepackage{cite}
\usepackage{amsmath,amssymb,amsfonts}
\usepackage{algorithmic}
\usepackage{graphicx}
\usepackage{textcomp}
\usepackage{xcolor}

\usepackage{booktabs}
\usepackage[table]{xcolor}
\usepackage{makecell}
\usepackage{graphicx} 

\definecolor{bestcol}{RGB}{255,200,200}    
\definecolor{secondcol}{RGB}{255,236,145}   
\newcommand{\best}[1]{\cellcolor{bestcol}#1}
\newcommand{\second}[1]{\cellcolor{secondcol}#1}

\newcommand{\metrichead}[1]{\makecell[c]{#1}}
\newcommand{\triphead}[1]{\multicolumn{3}{c}{#1}}
\newcommand{\trip}[3]{#1 & #2 & #3}
\newcommand{\twohead}[1]{\multicolumn{2}{c}{#1}}
\newcommand{\two}[2]{#1 & #2}
\newcommand{\fourhead}[1]{\multicolumn{4}{c}{#1}}
\newcommand{\four}[4]{#1 & #2 & #3 & #4}

\usepackage{mdframed}
\usepackage[most]{tcolorbox}
\usepackage{xcolor}

\usepackage{graphicx}
\usepackage{capt-of}

\title{\LARGE \bf DispFlow-GS: Displacement Flow Supervision with Motion Disentangling for Monocular Deformable 3D Gaussian Splatting
}

\author{
Thai Duy Nguyen$^{1}$, Haitian Zhang$^{1}$, and Addison Lin Wang$^{1,*}$%
\thanks{$^{1}$Nanyang Technological University, Singapore.               
{\tt\small \{nguyendu003,haitian003\}@e.ntu.edu.sg, linwang@ntu.edu.sg}}%
\thanks{$^{*}$Corresponding author.}%
}

\begin{document}

\maketitle
\thispagestyle{empty}
\pagestyle{empty}

\input{sections/abstract}
\input{sections/introduction}
\input{sections/related_works}
\input{sections/revisit_gflow}
\input{sections/methodology}
\input{sections/experiments}

\input{sections/failure_cases}
\input{sections/conclusion}

%
%
\bibliographystyle{ieeetr}
\bibliography{main}

\end{document}

%% file: sections/abstract.tex
\begin{abstract}
Accurate dynamic scene reconstruction is important for robotic perception, where temporally consistent representations of dynamic environments are essential. Deformable 3D Gaussian Splatting (3DGS) models dynamic scenes through deformation fields, and recent methods incorporate motion supervision by aligning rendered Gaussian flow with optical flow. However, we find that such Gaussian-flow-based supervision provides only limited improvements in motion modeling. We identify a fundamental limitation of this supervision paradigm, namely a domain gap between rendered Gaussian flow and optical flow. To address this limitation, we propose a motion supervision framework built on \textbf{Displacement Flow}, which splats per-Gaussian 3D displacements onto the image plane to provide direct and stable optimization signals. We further disentangle scene motion from camera motion via intermediate-view rendering, enabling more reliable motion priors and targeted constraints on deformation and geometry. We also observe a discrepancy between motion fidelity and image-based evaluation, where improved motion awareness does not necessarily translate into better rendered image quality or higher image-based metric scores. Motivated by this mismatch, we introduce Deformation-Rendering Consistency (\textbf{DRC}), a motion-aware metric that measures the alignment between predicted deformation and rendering improvement. Experiments on dynamic scene benchmarks show substantial improvements in motion localization and motion--rendering consistency, reaching up to 39\% and 6\%, respectively, while image-based metrics change by only about 0.1\%. These results confirm the observed mismatch between motion fidelity and image-based evaluation, demonstrating the significance of DRC for motion-aware evaluation. 
\end{abstract}

%% file: sections/introduction.tex
\section{Introduction}

Novel-view synthesis~\cite{kerbl20233d,mildenhall2021nerf,chen2022tensorf} and 3D reconstruction of dynamic scenes is important for applications such as Simultaneous Localization and Mapping (SLAM)~\cite{sun2025embracing}, and robotic navigation~\cite{wang2025dynorecon}. Building upon neural radiance field (NeRF)~\cite{mildenhall2021nerf} and 3D Gaussian Splatting (3DGS)~\cite{kerbl20233d}, recent dynamic 3DGS methods have demonstrated efficient reconstruction and rendering of time-varying scenes~\cite{wu20244d,yang2024deformable,li2024spacetime}.
Dynamic 3DGS approaches model temporal evolution through three main paradigms, including: explicit 4D primitive-based methods, deformation-field-based methods, and frame-wise training methods.  
Explicit 4D methods incorporate time directly into Gaussian representations~\cite{li2024spacetime,yang20244d}, while frame-wise methods optimize temporally varying Gaussian states~\cite{luiten2024dynamic}. In contrast, deformation-field-based 3DGS maintains a canonical Gaussian representation and learns a deformation field to predict time-dependent changes in Gaussian attributes~\cite{wu20244d,yang2024deformable}. This provides a compact and continuous representation of scene dynamics, with better parameter efficiency than explicit 4D primitives and stronger temporal coherence than independent frame-wise methods. In this work, we focus specifically on this deformation-field-based paradigm for motion-aware reconstruction.

\begin{figure}[t!]
  \centering
  \includegraphics[width=.99\linewidth]{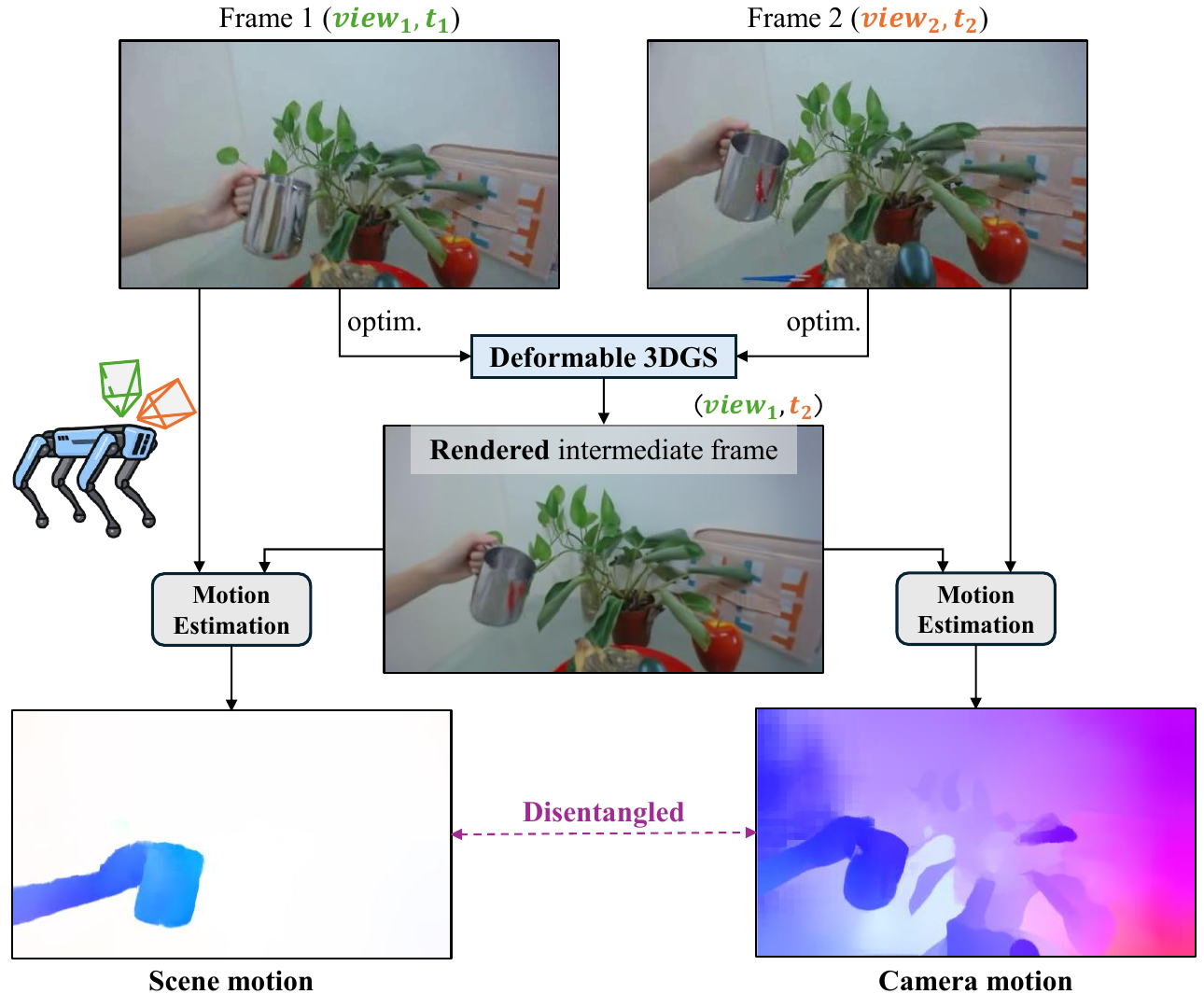}
  \vspace{-20pt}
  \captionof{figure}{
        \textbf{Highlight of our motion disentangling framework.}
        In dynamic scenarios, between frames there are two types of motion being entangled together: the camera motion (from $view_1$ to $view_2$), and the scene motion (from state at $t_1$ to $t_2$). For efficient motion supervision, such two motions need to be disentangle. Our method uses Deformable 3DGS to render an intermediate image at $(view_1,t_2)$, such that its motion relative to each input frame isolates a single motion component by keeping either the viewpoint or the scene state fixed during motion estimation.
    }
\vspace{-20pt}
  \label{fig:teaser_fig}
\end{figure}


Most deformable 3DGS methods rely primarily on photometric supervision between rendered and input images. However, photometric supervision is ambiguous for dynamic scenes, as different deformation patterns can produce similar appearance consistency. Recent works therefore introduce motion-aware constraints to improve deformation learning~\cite{gao2024gaussianflow,zhu2024motiongs,guo2024motion}. GaussianFlow~\cite{gao2024gaussianflow} renders the motion of projected Gaussians and supervises it using optical flow, while subsequent approaches further exploit Gaussian-derived motion for motion decomposition and deformation learning~\cite{zhu2024motiongs,lin2024gaussian}.
Our work focuses on this \textit{Gaussian-flow-based motion supervision} within deformation-field-based 3DGS. Despite different supervision targets and formulations, these approaches align motion rendered from deformed Gaussians with flow-derived motion cues. Our empirical analysis shows that such supervision provides only limited improvement in motion-aware deformation. We identify a key limitation: \textit{Gaussian flow and optical flow represent motion differently, creating a \textit{domain gap} between the rendered motion representation and its supervision target.}

To address this limitation, we propose a novel motion supervision framework using \textit{Displacement Flow}, obtained by splatting per-Gaussian 3D displacements onto the image plane. Rather than enforcing pixel correspondence between heterogeneous motion representations, Displacement Flow captures relative motion magnitude induced by Gaussian deformation. We further disentangle scene motion from camera-induced motion via intermediate-view rendering, as depicted in Fig.~\ref{fig:teaser_fig}, allowing motion priors to be estimated under a fixed viewpoint for more reliable supervision.
Since our focus is motion-aware deformation learning, image-based metrics such as PSNR, SSIM, and LPIPS cannot fully reflect motion fidelity. Empirical results in Sec.~\ref{sec:quant_results} show that improvements in motion quality do not necessarily translate into gains in these metrics. This distinction is particularly relevant to robotic perception, where correctly identifying dynamic regions can matter even when rendered appearance changes little. We therefore introduce \textit{Deformation--Rendering Consistency (DRC)}, which measures whether predicted deformation aligns with rendering improvement.

We evaluate our method on the NeRF-DS~\cite{yan2023nerf} and HyperNeRF~\cite{park2021hypernerf} benchmarks. Our approach consistently improves motion-aware metrics while image-based performance remains largely unchanged. We show that improved motion awareness does not necessarily translate into better rendered image quality or higher image-based metric scores, highlighting the need for motion-aware evaluation.

\noindent Our main contributions are summarized as follows:
\begin{enumerate}
    \item We identify a domain gap between rendered Gaussian flow and optical-flow supervision as a key limitation of motion learning in deformable 3DGS.
    \item We propose a motion supervision framework using \textit{Displacement Flow} and scene-motion decomposition for more efficient deformation learning.
    \item We reveal a discrepancy between motion fidelity and image-based evaluation, motivating \textit{Deformation--Rendering Consistency} for motion-aware evaluation.
\end{enumerate}

%% file: sections/related_works.tex
\section{Related Works}

\noindent \textbf{Dynamic Scene Reconstruction.}
The NeRF framework~\cite{mildenhall2021nerf} has been extended to dynamic scenes through deformation fields that map observations to a shared canonical space~\cite{park2021nerfies,pumarola2021d}, direct spatiotemporal modeling~\cite{li2023dynibar,xian2021space,gao2021dynamic}, and structured representations~\cite{cao2023hexplane,fridovich2023k,shao2023tensor4d}. Similarly, many works explore dynamic scene reconstruction with 3DGS~\cite{katsumata2024compact,lei2025mosca}. Dynamic 3DGS methods can be broadly grouped into explicit 4D primitive-based, deformation-field-based, and frame-wise approaches. Explicit 4D methods incorporate time directly into Gaussian primitives~\cite{li2024spacetime,sun2025splatflow}, while frame-wise methods optimize Gaussian states across timestamps~\cite{luiten2024dynamic,sun20243dgstream}. In contrast, deformation-field-based methods maintain canonical Gaussians and predict time-dependent transformations through a learned deformation field~\cite{wu20244d,yang2024deformable}. Our work focuses on the deformation-field-based paradigm and studies \textit{motion supervision for more reliable motion-aware deformation learning.}

\smallskip
\noindent \textbf{Motion-Aware Deformable 3DGS.}
Recent methods introduce motion supervision to improve deformation learning in deformable 3DGS. GaussianFlow~\cite{gao2024gaussianflow} supervises rendered Gaussian flow using optical flow, while MotionGS~\cite{zhu2024motiongs} further decomposes camera and scene motion before applying flow guidance. Guo et al.~\cite{guo2024motion} incorporate uncertainty-aware flow supervision, whereas Xie et al.~\cite{xie2024gaussian} estimate Gaussian motion using a differentiable Lucas--Kanade formulation without explicit optical-flow supervision. Despite different designs, these approaches still align Gaussian-derived motion with flow-based targets whose motion definitions are inherently different, introducing the domain gap studied in this work.
In contrast, we supervise \textit{Displacement Flow}, constructed from per-Gaussian 3D displacements, using directly estimated scene motion under a fixed viewpoint, avoiding direct correspondence matching between Gaussian-derived motion and optical flow, which represent motion differently.

\begin{figure}[t!]
  \centering
  \includegraphics[width=0.85\linewidth]{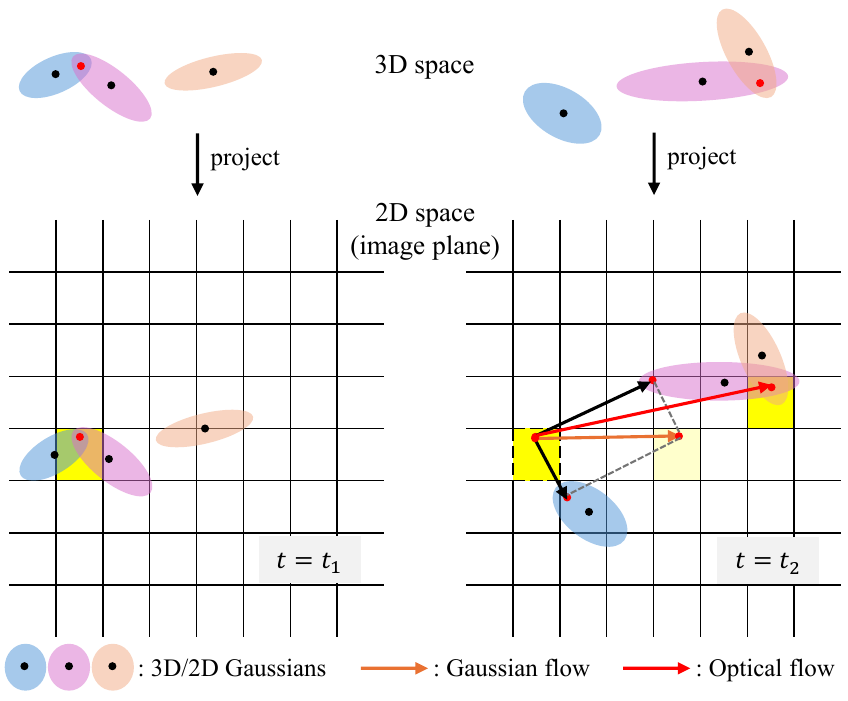}
    \vspace{-10pt}
  \caption{\textbf{Gaussian flow formulation and its discrepancy with optical flow.}
Gaussian flow is the $\alpha$-composited weighted sum of pixel displacements induced by projected 2D Gaussians. It considers only the Gaussian composition at $t_1$, ignoring changes in contributing Gaussians and blending weights due to deformation and viewpoint variation. In contrast, optical flow uses pixel positions at both $t_1$ and $t_2$, leading to a domain gap between two representations. 
}
\vspace{-20pt}
  \label{fig:gaussian_flow}
\end{figure}

\smallskip
\noindent \textbf{Motion Disentanglement for Robotics.}
Robotic perception in dynamic environments requires distinguishing platform-induced motion from motion intrinsic to the scene, as both are entangled in temporal observations. Recent works address this through dynamic--static scene decomposition for autonomous driving~\cite{doll2024dualad}, scene-flow-based motion reasoning~\cite{kim2025flow4d,khoche2025ssf}, and motion segmentation using scene-flow and ego-motion consistency with 4D radar~\cite{liu2025self} or temporal LiDAR~\cite{yi2025moving}. Image-based methods further combine optical flow and geometric constraints to separate scene motion from camera-induced motion~\cite{goli2025romo}. When camera-induced motion and scene motion are not separated, the observed motion is difficult to attribute reliably to actual scene dynamics. Such ambiguity also affects dynamic scene representations for robotic perception, where inter-frame motion jointly reflects viewpoint change and scene dynamics. \textit{We address it through intermediate-view rendering, fixing the viewpoint to isolate scene motion for more reliable deformation supervision.}

\begin{figure*}[t!]
  \centering
  \includegraphics[width=0.82\textwidth]{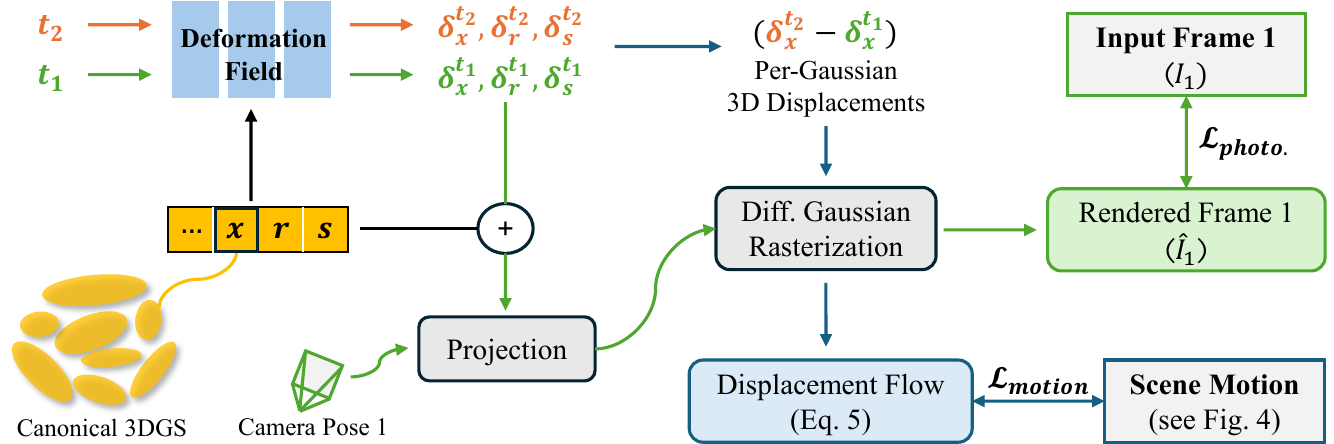}
    \vspace{-6pt}
  \caption{\textbf{Overview of the proposed DispFlow-GS framework.} The optimization consists of two components: photometric loss and motion loss. The photometric loss is the standard rendered image loss. The motion loss is defined by our Displacement Flow Supervision, where displacement flow is obtained by splatting per-Gaussian 3D displacements onto the image plane and supervising them against scene motion.
  }
  \vspace{-15pt}
  \label{fig:methodology}
\end{figure*}

%% file: sections/revisit_gflow.tex
\section{Revisit The Motion Supervision for Deformable 3DGS}

This section revisits a widely adopted motion supervision scheme in existing works \cite{gao2024gaussianflow, zhu2024motiongs, xie2024gaussian, guo2024motion} that aims to improve motion awareness. In this approach, Gaussian flow is rendered by compositing the motion of projected 2D Gaussians through $\alpha$-blending. The rendered flow map is then aligned with target optical flow to provide supervision. We then examine whether these two representations encode motion consistently, revealing their domain mismatch.

\subsection{Formulation of Gaussian Flow Supervision}

Motion supervision in deformable 3DGS typically extracts a motion representation from deformed Gaussians and supervises it using priors from off-the-shelf flow models. Despite different supervision targets and losses, many methods~\cite{gao2024gaussianflow,zhu2024motiongs,xie2024gaussian,guo2024motion} adopt Gaussian flow as the motion representation. Since the supervision directly compares this rendered motion with an external flow prior, its formulation determines how effectively the target motion constrains the deformation field.

Introduced in GaussianFlow~\cite{gao2024gaussianflow}, Gaussian flow $F^{G}_{t_1 \rightarrow t_2}$ is the $\alpha$-composited displacement of projected 2D Gaussians:
\begin{equation}
\label{eq:gaussian_flow}
\begin{split}
F^{G}_{t_1 \rightarrow t_2}
&= \sum_{i=1}^{K} w_i
\left( \mathbf{x}_{i,t_2} - \mathbf{x}_{t_1} \right) \\
&= \sum_{i=1}^{K} w_i
\left(
\Sigma_{i,t_2}\Sigma_{i,t_1}^{-1}
(\mathbf{x}_{t_1}-\mu_{i,t_1})
+\mu_{i,t_2}-\mathbf{x}_{t_1}
\right).
\end{split}
\end{equation}
Here, $K$ is the number of contributing Gaussians and
$w_i=\frac{T_i\alpha_i}{\sum_jT_j\alpha_j}$ is the normalized $\alpha$-blending weight. $\mathbf{x}_{t_1}$ and $\mathbf{x}_{i,t_2}$ denote the pixel coordinates before and after Gaussian deformation, while $\mu_{i,t}$ and $\Sigma_{i,t}$ are the Gaussian mean and covariance. The computation is illustrated in Fig.~\ref{fig:gaussian_flow}.
Applying \eqref{eq:gaussian_flow} to all pixels produces the rendered Gaussian flow, supervised by pseudo ground-truth optical flow:
\begin{equation}
\label{eq:flow_loss}
\mathcal{L}_{\text{flow}}
=
\left\|
F^{o}_{t_1 \rightarrow t_2}(\mathbf{x}_{t_1})
-
F^{G}_{t_1 \rightarrow t_2}
\right\|,
\end{equation}
where $F^{o}_{t_1 \rightarrow t_2}$ is estimated using a pre-trained flow model.

\subsection{Problems of Gaussian Flow Supervision Scheme}
\label{sec:problems}

From \eqref{eq:gaussian_flow}, Gaussian flow at a pixel is computed using the composition weights $w_i$ at time $t_1$, modeling pixel displacement as the weighted aggregation of the displacements of Gaussians contributing at $t_1$. This formulation \textit{implicitly assumes that Gaussian contributions to a pixel remain consistent across timestamps}. In practice, however, both the contributing Gaussians and their $\alpha$-blending weights change over time due to deformation and reprojection, leading to a mismatch between the supervision signal and the rendered motion representation used for deformation learning.

Consequently, Gaussian flow does not model explicit correspondence between pixel locations at $t_1$ and $t_2$, but instead represents accumulated motion induced by Gaussian deformation at $t_1$. In contrast, optical flow establishes a direct mapping between pixel positions at $t_1$ and their landing positions at $t_2$, as illustrated in Fig.~\ref{fig:gaussian_flow}. This difference introduces a domain gap between Gaussian flow and optical flow, preventing the semantic meaning of optical flow from being faithfully captured under this formulation. 



\begin{figure*}[t!]
  \centering
  \includegraphics[width=0.85\textwidth]{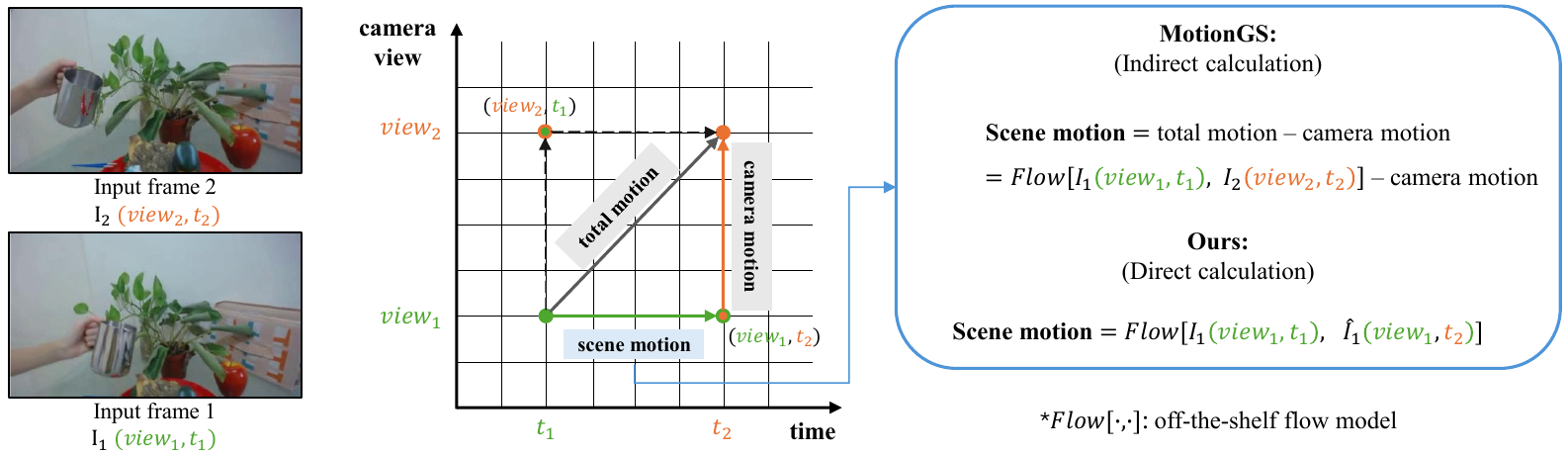}
    \vspace{-10pt}
  \caption{\textbf{Illustration of scene motion calculation.}
For an input pair $(view_1,t_1)$ and $(view_2,t_2)$, total motion includes scene and camera motion. 
MotionGS estimates cross-view flow and subtracts camera motion, which is error-prone due to large cross-view displacements. 
Instead, we render an intermediate frame at $(view_1,t_2)$ and compute flow between $(view_1,t_1)$ and $(view_1,t_2)$ to isolate scene motion under a fixed viewpoint.}
\vspace{-8pt}
  \label{fig:motion_decompose}
\end{figure*}

%% file: sections/methodology.tex
\section{Methodology}

\subsection{Displacement Flow Supervision}
\label{sec:Lmotion}
\noindent \textbf{Design Rationales.}
A natural way to address the domain gap in Sec.~\ref{sec:problems} is to model Gaussian composition at both $t_1$ and $t_2$ and establish cross-time correspondence. However, obtaining correspondence at $t_2$ would already amount to solving optical flow, making this formulation circular. We therefore reinterpret Gaussian-derived motion as a motion magnitude map rather than a correspondence field. This motivates a representation that captures relative Gaussian motion without requiring explicit cross-time pixel correspondence.

\smallskip
\noindent \textbf{Formulation of Displacement Flow.}
We propose \emph{Displacement Flow}, derived by splatting per-Gaussian 3D displacements onto the image plane (see Fig.~\ref{fig:methodology}).
At each step, a pair of images at timestamps $t_1$ and $t_2$ is processed, producing two sets of deformations from the canonical 3DGS:
\begin{equation}
    \delta_{x_i}^{t_1}, \delta_{r_i}^{t_1}, \delta_{s_i}^{t_1} = \mathcal{D}(x_i, t_1), 
    \quad 
    \delta_{x_i}^{t_2}, \delta_{r_i}^{t_2}, \delta_{s_i}^{t_2} = \mathcal{D}(x_i, t_2).
\end{equation}

The 3D displacement of each Gaussian from $t_1$ to $t_2$ is computed as
\begin{equation}
    \delta_{x_i}^{t_1t_2} = \delta_{x_i}^{t_2} - \delta_{x_i}^{t_1}.
\end{equation}

Displacement Flow is then obtained by splatting these 3D displacements onto the image plane using $\alpha$-blending:
\begin{equation}
F_{\text{disp}} = \sum_{i=1}^{K} w_i \, \delta_{x_i}^{t_1t_2},
\end{equation}
where $w_i$ is the $\alpha$-blending weight and $K$ is the number of contributing Gaussians. Unlike Gaussian flow, Displacement Flow is constructed directly from 3D center displacements, providing a direct gradient path to the deformation field. 

\smallskip
\noindent \textbf{Supervision.}
The displacement flow $F_{\text{disp}}$ is supervised against scene motion $F_{\text{scene}}$, which excludes camera-induced motion (detailed in Sec.~\ref{sec:disentangle}). Following the discussion above, both flows are normalized into a motion magnitude representation in $[0,1]$ using min-max normalization.

The motion loss is defined as
\begin{equation}
\mathcal{L}_{\text{motion}} =
\left\|
\hat{F}_{\text{scene}} - \hat{F}_{\text{disp}}
\right\|_1.
\end{equation}

Rather than enforcing exact pixel correspondence, this supervision aligns relative motion structures across the scene, encouraging deformation updates that are consistent with dynamic regions while mitigating the domain gap limitation discussed in Sec.~\ref{sec:problems}.

\subsection{Motion Disentangling via Intermediate Image Rendering}
\label{sec:disentangle}

At each optimization step, we process an image pair from different viewpoints and timestamps. Two motion components occur simultaneously: \emph{scene motion} (temporal deformation) and \emph{camera motion} (viewpoint change), denoted as $(view_1, t_1) \to (view_2, t_2)$. Since Displacement Flow models only scene motion, these components must be disentangled.
Using the differentiable rendering of deformable 3DGS, we render an intermediate state at $(view_1, t_2)$, decomposing the transformation as:
\begin{equation}
(view_1,t_1) \rightarrow (view_1,t_2) \rightarrow (view_2,t_2),
\end{equation}
where the first stage isolates scene motion and the second camera motion. Rendering at $(view_2, t_1)$ produces the reverse order. We adopt the former so scene motion is expressed under $view_1$, which is directly constrained by photometric loss, leading to more stable supervision.

These decomposed pairs enable separate estimation of scene motion flow and camera flow using off-the-shelf optical flow models. The resulting flows provide pseudo ground-truth motion priors during optimization. This decomposition also improves flow reliability, as directly estimating flow between the original pair can be inaccurate under large or complex motion. See Fig.~\ref{fig:motion_decompose} for illustration.

\subsection{Optimization}

The overall training objective is
\begin{equation}
\mathcal{L}
=
\mathcal{L}_{\text{photo}}
+
\alpha_t\, \mathcal{L}_{\text{motion}},
\end{equation}
where $\mathcal{L}_{\text{photo}}$ is the photometric loss and $\mathcal{L}_{\text{motion}}$ is the proposed motion supervision term. The coefficient $\alpha_t$ controls the contribution of motion supervision during training.

To stabilize optimization, motion supervision is introduced after an initial warm-up period and its weight is gradually increased. For iterations $t \ge t_0$, $\alpha_t$ is linearly ramped from $0.1\alpha$ to $\alpha$, while $\alpha_t = 0$ for $t < t_0$. This scheduling prevents motion supervision from dominating early training before photometric reconstruction stabilizes, allowing the deformation field to first learn a reasonable geometric initialization and later refine motion-consistent deformation.

%% file: sections/experiments.tex
\section{Experiments}
\label{sec:experiments}

\noindent \textbf{Implementation Details.}
Training runs for 20{,}000 iterations, with the first 3{,}000 iterations used for warm-up. Motion supervision is introduced at iteration 10{,}000 with a final weight of $\alpha = 0.1$ for both the NeRF-DS~\cite{yan2023nerf} and HyperNeRF~\cite{park2021hypernerf} datasets. The spatial learning-rate scale for the deformation field is set to 12. We use the pre-trained GMFlow~\cite{xu2022gmflow} model to estimate scene motion. Other hyperparameters follow the baseline method~\cite{yang2024deformable}. Experiments are conducted using PyTorch~\cite{paszke2019pytorch} on a single NVIDIA RTX 5090 GPU.

\begin{table*}[t!]
  \centering
  \caption{Quantitative comparison using \textbf{image-based metrics} on the \textbf{NeRF-DS} dataset per scene. \\
  We highlight the \colorbox{bestcol}{best} and \colorbox{secondcol}{second-best} results in each scene.}
  \vspace{-8pt}
  \label{tab:nerfds_photo}
  \resizebox{0.85\linewidth}{!}{%
  \begin{tabular}{l*{12}{c}}
    \toprule
    & \triphead{\textbf{Sieve}} & \triphead{\textbf{Plate}} &
      \triphead{\textbf{Bell}} & \triphead{\textbf{Press}} \\
    \cmidrule(lr){2-4}\cmidrule(lr){5-7}\cmidrule(lr){8-10}\cmidrule(lr){11-13}
    \textbf{Method} &
      \metrichead{PSNR\,$\uparrow$} & \metrichead{SSIM\,$\uparrow$} & \metrichead{LPIPS\,$\downarrow$} & 
      \metrichead{PSNR\,$\uparrow$} & \metrichead{SSIM\,$\uparrow$} & \metrichead{LPIPS\,$\downarrow$} & 
      \metrichead{PSNR\,$\uparrow$} & \metrichead{SSIM\,$\uparrow$} & \metrichead{LPIPS\,$\downarrow$} & 
      \metrichead{PSNR\,$\uparrow$} & \metrichead{SSIM\,$\uparrow$} & \metrichead{LPIPS\,$\downarrow$}  \\
    \midrule
    3D\mbox{-}GS~\cite{kerbl20233d}        & \trip{23.16}{0.8203}{0.2247} & \trip{16.14}{0.6970}{0.4093} & \trip{21.01}{0.7885}{0.2503} & \trip{22.89}{0.8163}{0.2904} \\
    TiNeuVox~\cite{fang2022fast}        & \trip{21.49}{0.8265}{0.3176} & \trip{\best{20.58}}{0.8027}{0.3317} & \trip{23.08}{0.8242}{0.2568} & \trip{24.47}{0.8613}{0.3001} \\
    HyperNeRF~\cite{park2021hypernerf}      & \trip{25.43}{\second{0.8798}}{0.1645} & \trip{18.93}{0.7709}{0.2940} & \trip{23.06}{0.8097}{0.2052} & \trip{\best{26.15}}{\best{0.8897}}{0.1959} \\
    NeRF\mbox{-}DS~\cite{yan2023nerf}    & \trip{\best{25.78}}{\best{0.8900}}{\best{0.1472}} & \trip{\second{20.54}}{0.8042}{\best{0.1996}} & \trip{23.19}{0.8212}{0.1867} & \trip{\second{25.72}}{0.8618}{0.2047} \\
    Deformable\mbox{-}3DGS~\cite{yang2024deformable} & \trip{25.27}{0.8682}{\second{0.1532}} & \trip{20.51}{\best{0.8090}}{\second{0.2271}} & \trip{\second{25.07}}{\second{0.8446}}{\second{0.1725}} & \trip{25.45}{0.8619}{\second{0.1941}} \\
    MotionGS~\cite{zhu2024motiongs}  & \trip{\second{25.49}}{0.8443}{0.2193} & \trip{20.19}{0.7920}{0.2685} & \trip{24.89}{0.8198}{0.2465} & \trip{25.62}{0.8539}{0.2415} \\
    DispFlow-GS     & \trip{25.33}{0.8685}{0.1570} & \trip{20.27}{\second{0.8045}}{0.2344} & \trip{\best{25.19}}{\best{0.8459}}{\best{0.1676}} & \trip{25.56}{\second{0.8644}}{\best{0.1940}} \\
    \midrule
    & \triphead{\textbf{Cup}} & \triphead{\textbf{As}} &
      \triphead{\textbf{Basin}} & \triphead{\textbf{\textit{Mean}}} \\
    \cmidrule(lr){2-4}\cmidrule(lr){5-7}\cmidrule(lr){8-10}\cmidrule(lr){11-13}
    \textbf{Method} &
      \metrichead{PSNR\,$\uparrow$} & \metrichead{SSIM\,$\uparrow$} & \metrichead{LPIPS\,$\downarrow$} & 
      \metrichead{PSNR\,$\uparrow$} & \metrichead{SSIM\,$\uparrow$} & \metrichead{LPIPS\,$\downarrow$} & 
      \metrichead{PSNR\,$\uparrow$} & \metrichead{SSIM\,$\uparrow$} & \metrichead{LPIPS\,$\downarrow$} & 
      \metrichead{PSNR\,$\uparrow$} & \metrichead{SSIM\,$\uparrow$} & \metrichead{LPIPS\,$\downarrow$}  \\
    \midrule
    3D\mbox{-}GS~\cite{kerbl20233d}        & \trip{21.71}{0.8304}{0.2548} & \trip{22.69}{0.8017}{0.2994} & \trip{18.42}{0.7170}{0.3153} & \trip{20.86}{0.7816}{0.2920} \\
    TiNeuVox~\cite{fang2022fast}        & \trip{19.71}{0.8109}{0.3643} & \trip{21.26}{0.8289}{0.3967} & \trip{\best{20.66}}{0.8145}{0.2690} & \trip{21.61}{0.8241}{0.3195} \\
    HyperNeRF~\cite{park2021hypernerf}      & \trip{\second{24.59}}{0.8770}{\second{0.1650}} & \trip{25.58}{\best{0.8949}}{\second{0.1777}} & \trip{\second{20.41}}{\best{0.8199}}{\second{0.1911}} & \trip{23.45}{\second{0.8488}}{0.1991} \\
    NeRF\mbox{-}DS~\cite{yan2023nerf}    & \trip{\best{24.91}}{0.8741}{0.1737} & \trip{25.13}{0.8778}{\best{0.1741}} & \trip{19.96}{\second{0.8166}}{\best{0.1855}} & \trip{23.60}{\best{0.8494}}{\best{0.1816}} \\
    Deformable\mbox{-}3DGS~\cite{yang2024deformable} & \trip{24.20}{\second{0.8816}}{0.1681} & \trip{\best{26.12}}{\second{0.8814}}{0.1856} & \trip{19.68}{0.7927}{0.1918} & \trip{\second{23.76}}{0.8485}{\second{0.1846}} \\
    MotionGS ~\cite{zhu2024motiongs}   & \trip{24.23}{0.8553}{0.2109} & \trip{26.02}{0.8626}{0.2256} & \trip{19.79}{0.7751}{0.2426} & \trip{23.75}{0.8290}{0.2364} \\
    DispFlow-GS  & \trip{24.47}{\best{0.8876}}{\best{0.1620}} & \trip{\second{26.08}}{0.8770}{0.1991} & \trip{19.68}{0.7914}{0.1941} & \trip{\best{23.80}}{0.8485}{0.1869} \\
    \bottomrule
  \end{tabular}}
  \vspace{-5pt}
\end{table*}

\smallskip
\noindent \textbf{Benchmark Datasets.} We evaluate our method on two representative real-world dynamic scene benchmarks: NeRF-DS~\cite{yan2023nerf} and HyperNeRF~\cite{park2021hypernerf}. The image resolutions and train–test splits follow the original dataset protocols.

\subsection{Evaluation Metrics}

\noindent \textbf{Image-based Metrics.}
Following prior works, we report PSNR, SSIM, and LPIPS for image reconstruction quality.

\smallskip
\noindent \textbf{Motion-based Metrics.}
Image-based metrics mainly measure appearance and do not directly assess motion quality. We therefore introduce Deformation-Rendering Consistency (DRC) to measure whether predicted deformation aligns with rendering improvement. To avoid bias from evaluating solely with our proposed DRC, we also use MMAP, which follows the standard Average Precision (AP) evaluation protocol.

\paragraph{Deformation-Rendering Consistency (DRC)}
DRC measures whether deformation-induced motion improves reconstruction. It does not aim to measure true motion correspondence, but rather whether predicted motion is concentrated in regions where deformation reduces rendering error.

For consecutive frames $t{-}1$ and $t$, let $I_{\mathrm{frozen}}$ denote the render using the previous deformation field $\mathbf{\mathcal{D}}(t{-}1)$ and $I_{\mathrm{full}}$ the render using the updated deformation $\mathbf{\mathcal{D}}(t)$. The deformation gain is
\begin{equation}
\label{eq:drc_gain}
\Delta\mathcal{E}(p)
=
|I_{\mathrm{frozen}}(p)-I_{\mathrm{gt}}(p)|
-
|I_{\mathrm{full}}(p)-I_{\mathrm{gt}}(p)|,
\end{equation}
where $\Delta\mathcal{E}(p)>0$ indicates improved reconstruction. Let $\hat{F}_{\mathrm{disp}}(p)\in[0,1]$ denote the normalized displacement flow. DRC is defined as
\begin{equation}
\label{eq:drc}
\mathrm{DRC}
=
\frac{
\sum_{p}\hat{F}_{\mathrm{disp}}(p)\,
\mathbf{1}[\Delta\mathcal{E}(p)>0]
}{
\sum_{p}\hat{F}_{\mathrm{disp}}(p)+\varepsilon
}
\in[0,1],
\end{equation}
where higher DRC indicates stronger alignment between predicted motion and error-reducing deformation.

\smallskip
\paragraph{Motion Mask Average Precision (MMAP)}
MMAP evaluates motion localization using the standard AP evaluation protocol. Let $M_f(p)\!\in\!\{0,1\}$ denote the ground-truth motion label and $\hat{F}_{\mathrm{disp},f}(p)\!\in\![0,1]$ the normalized displacement flow as per-pixel motion confidence. For each frame, AP is computed from the precision--recall curve by thresholding $\hat{F}_{\mathrm{disp},f}$ against $M_f$, and MMAP is the mean AP over all test frames. Higher MMAP indicates better alignment between predicted motion and true dynamic regions.
Both metrics require displacement flow and are evaluated on image pairs, consistent with the training procedure.

\begin{table}[t!]
  \centering
  \caption{Quantitative comparison using \textbf{motion-based metrics} on the \textbf{NeRF-DS} dataset per scene. 
  }
  \vspace{-8pt}
  \label{tab:nerfds_motion}
  \resizebox{\linewidth}{!}{%
  \begin{tabular}{l*{8}{c}}
    \toprule
    & \twohead{\textbf{Sieve}} & \twohead{\textbf{Plate}} &
      \twohead{\textbf{Bell}} & \twohead{\textbf{Press}} \\
    \cmidrule(lr){2-3}\cmidrule(lr){4-5}\cmidrule(lr){6-7}\cmidrule(lr){8-9}
    \textbf{Method} &
      \metrichead{MMAP\,$\uparrow$} & \metrichead{DRC\,$\uparrow$} & \metrichead{MMAP\,$\uparrow$} & \metrichead{DRC\,$\uparrow$} &
      \metrichead{MMAP\,$\uparrow$} & \metrichead{DRC\,$\uparrow$} & \metrichead{MMAP\,$\uparrow$} & \metrichead{DRC\,$\uparrow$} \\
    \midrule
    Deformable\mbox{-}3DGS~\cite{yang2024deformable} & \two{0.4220}{0.6320} & \two{\second{0.7514}}{0.7088} & \two{\second{0.8562}}{\second{0.7608}} & \two{0.7346}{\second{0.6994}} \\
    MotionGS~\cite{zhu2024motiongs} & \two{\second{0.5905}}{\second{0.6600}} & \two{0.7149}{\second{0.7095}} & \two{0.8236}{0.7395} & \two{\second{0.7398}}{0.6867} \\
    DispFlow-GS & \two{\best{0.8724}}{\best{0.6957}} & \two{\best{0.9204}}{\best{0.7573}} & \two{\best{0.8936}}{\best{0.7769}} & \two{\best{0.9128}}{\best{0.7133}} \\
    \midrule
    & \twohead{\textbf{Cup}} & \twohead{\textbf{As}} &
      \twohead{\textbf{Basin}} & \twohead{\textbf{\textit{Mean}}} \\
    \cmidrule(lr){2-3}\cmidrule(lr){4-5}\cmidrule(lr){6-7}\cmidrule(lr){8-9}
    \textbf{Method} &
      \metrichead{MMAP\,$\uparrow$} & \metrichead{DRC\,$\uparrow$} & \metrichead{MMAP\,$\uparrow$} & \metrichead{DRC\,$\uparrow$} &
      \metrichead{MMAP\,$\uparrow$} & \metrichead{DRC\,$\uparrow$} & \metrichead{MMAP\,$\uparrow$} & \metrichead{DRC\,$\uparrow$} \\
    \midrule
    Deformable\mbox{-}3DGS~\cite{yang2024deformable} & \two{0.6066}{0.6060} & \two{0.3566}{0.6422} & \two{0.8205}{0.6127} & \two{0.6497}{0.6660} \\
    MotionGS~\cite{zhu2024motiongs} & \two{\second{0.6733}}{\second{0.6132}} & \two{\second{0.3818}}{\second{0.6516}} & \two{\second{0.8538}}{\second{0.6193}} & \two{\second{0.6825}}{\second{0.6685}} \\
    DispFlow-GS & \two{\best{0.9356}}{\best{0.7036}} & \two{\best{0.8449}}{\best{0.6868}} & \two{\best{0.9308}}{\best{0.6231}} & \two{\best{0.9015}}{\best{0.7081}} \\
    
    \bottomrule
  \end{tabular}}
  \vspace{-15pt}
\end{table}

\begin{table*}[t!]
  \centering
  \caption{Quantitative comparison on the \textbf{HyperNeRF} dataset per scene. 
  }
  \vspace{-8pt}
  \label{tab:hypernerf_quanti}
  \resizebox{0.85\linewidth}{!}{%
  \begin{tabular}{l*{12}{c}}
    \toprule
    & \fourhead{\textbf{Chicken}} & \fourhead{\textbf{Banana}} &
      \fourhead{\textbf{Mean}}  \\
    \cmidrule(lr){2-5}\cmidrule(lr){6-9}\cmidrule(lr){10-13}
    \textbf{Method} &
      \metrichead{PSNR\,$\uparrow$} & \metrichead{SSIM\,$\uparrow$} & \metrichead{MMAP\,$\uparrow$} & \metrichead{DRC\,$\uparrow$} & 
      \metrichead{PSNR\,$\uparrow$} & \metrichead{SSIM\,$\uparrow$} & \metrichead{MMAP\,$\uparrow$} & \metrichead{DRC\,$\uparrow$} &
      \metrichead{PSNR\,$\uparrow$} & \metrichead{SSIM\,$\uparrow$} & \metrichead{MMAP\,$\uparrow$} & \metrichead{DRC\,$\uparrow$} \\
    \midrule

    HyperNeRF~\cite{park2021hypernerf}
    & \four{\second{27.40}}{0.630}{--}{--}
    & \four{22.10}{0.720}{--}{--}
    & \four{\second{24.80}}{0.680}{--}{--} \\
    
    TiNeuVox~\cite{fang2022fast}
    & \four{\best{28.20}}{\best{0.790}}{--}{--}
    & \four{24.40}{0.640}{--}{--}
    & \four{\best{26.30}}{0.720}{--}{--} \\
    
    Deformable\mbox{-}3DGS~\cite{yang2024deformable}
    & \four{23.58}{0.646}{0.529}{\second{0.729}}
    & \four{\best{25.16}}{\best{0.831}}{0.749}{\best{0.760}}
    & \four{24.37}{\second{0.738}}{0.639}{\second{0.744}} \\
    
    MotionGS~\cite{zhu2024motiongs}
    & \four{23.44}{0.638}{\second{0.625}}{\second{0.729}}
    & \four{24.90}{0.829}{\second{0.751}}{0.752}
    & \four{24.17}{0.733}{\second{0.688}}{0.741} \\
    
    DispFlow-GS
    & \four{23.66}{\second{0.649}}{\best{0.664}}{\best{0.743}}
    & \four{\second{24.97}}{\second{0.830}}{\best{0.774}}{\second{0.758}}
    & \four{24.31}{\best{0.739}}{\best{0.719}}{\best{0.750}} \\
    \bottomrule
  \end{tabular}}
  \vspace{-10pt}
\end{table*}

\subsection{Results}
\label{sec:quant_results}
\subsubsection{Results on the NeRF-DS dataset}



\begin{figure}[t!]
  \centering
  \includegraphics[width=\linewidth]{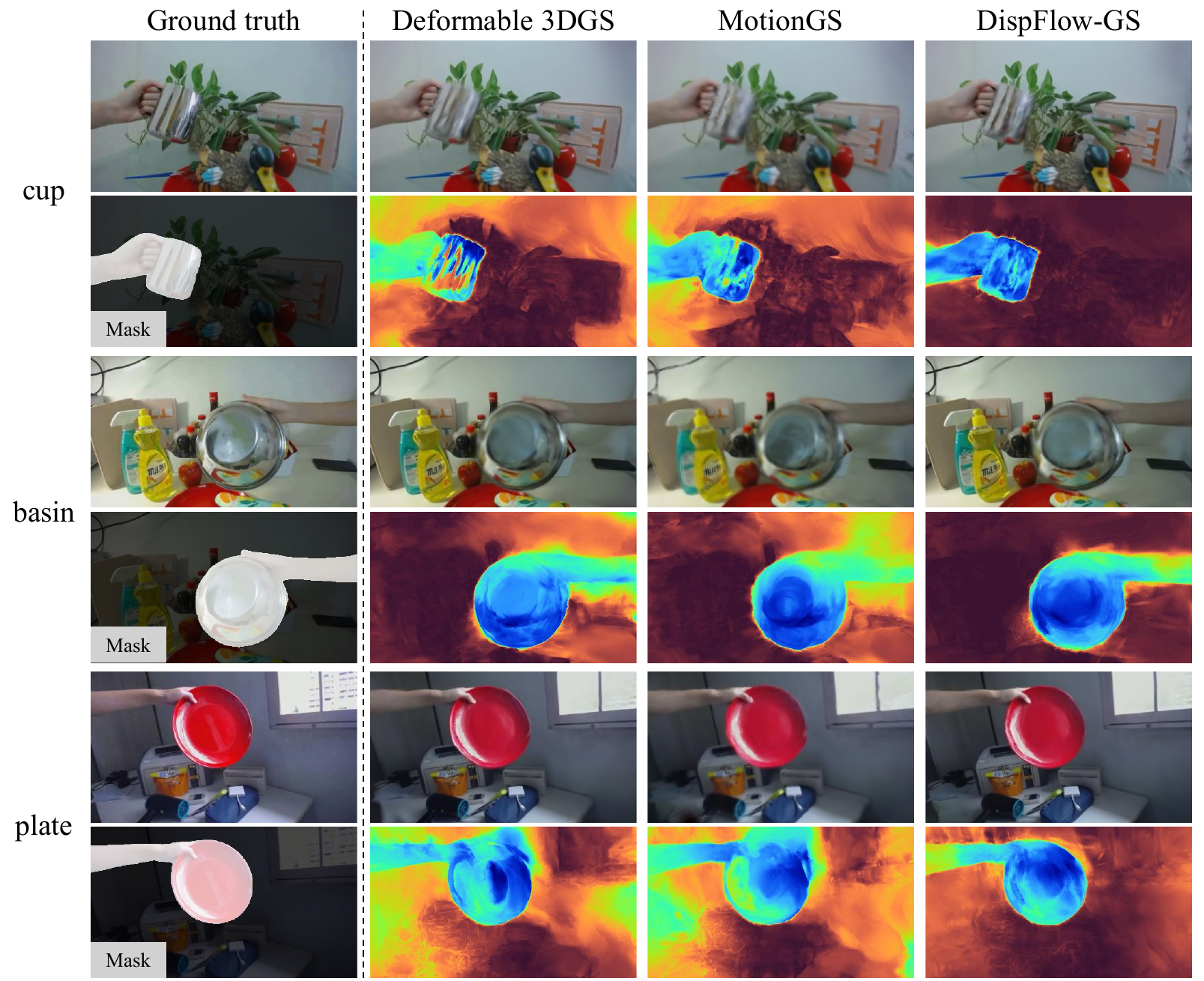}
  \vspace{-20pt}
  \caption{\textbf{Qualitative results on representative scenes of NeRF-DS dataset.}
  For each scene, the top row shows the rendered images, while the bottom row shows the corresponding Displacement Flow.
  }
  \vspace{-15pt}
  \label{fig:nerfds_quali}
\end{figure}

Quantitative results are reported in Tab.~\ref{tab:nerfds_photo} and Tab.~\ref{tab:nerfds_motion}. \textbf{DispFlow-GS achieves the best motion-based performance across all scenes}, improving mean MMAP and DRC over the photometric-only Deformable 3DGS~\cite{yang2024deformable} by approximately \textbf{39\%} and \textbf{6\%}, respectively, while PSNR changes by only about 0.1\%. This indicates that the improvement brought by motion supervision is expressed primarily in the learned deformation rather than in conventional reconstruction quality. Qualitative results in Fig.~\ref{fig:nerfds_quali} support this observation, where our method localizes deformation more consistently to dynamic regions and suppresses responses in static backgrounds. The \emph{basin} scene provides a particularly clear example: MotionGS produces visibly blurrier reconstructions while attaining the highest PSNR among the three deformable-3DGS methods, showing that a better image-based score does not necessarily correspond to more faithful motion modeling.

This discrepancy reflects the different objectives of photometric and motion supervision. Photometric optimization can improve appearance by allowing flexible deformation that explains the observed images, without requiring the resulting deformation to correspond to the actual dynamic regions. In contrast, motion supervision explicitly constrains where and how deformation should occur, which may restrict this photometric flexibility while producing more meaningful scene motion. \textbf{Thus, substantial improvements in motion-aware deformation can occur with little change in image-based metrics}, motivating the use of motion-specific metrics when evaluating motion-supervised dynamic reconstruction.

\subsubsection{Results on the HyperNeRF dataset}



\begin{figure}[t!]
  \centering
 \includegraphics[width=0.8\linewidth]{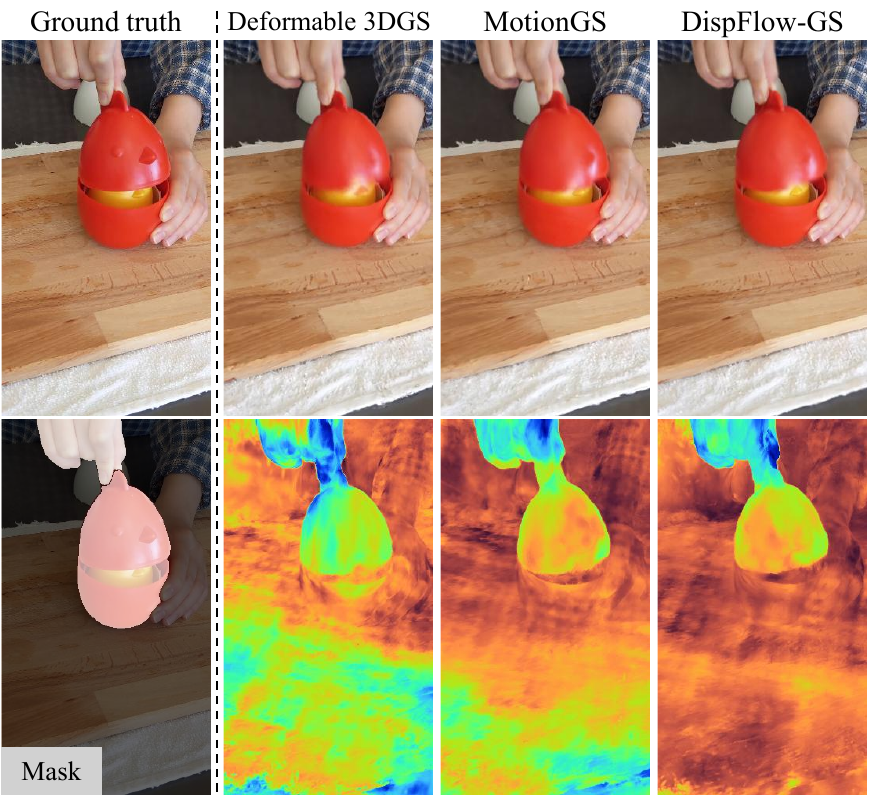}
 \vspace{-8pt}
  \caption{\textbf{Qualitative results on `chicken' scene of the HyperNeRF dataset.}
}
  \label{fig:hypernerf_quali}
\end{figure}

We report results on the \textit{``chicken''} and \textit{``banana''} scenes in Tab.~\ref{tab:hypernerf_quanti}, with qualitative comparisons in Fig.~\ref{fig:hypernerf_quali} for the \textit{``chicken''} scene. We omit the remaining scenes because several HyperNeRF sequences contain noticeable camera-pose inaccuracies, as also reported in Deformable 3DGS~\cite{yang2024deformable}. Such errors can degrade deformation learning in deformable 3DGS methods and also affect intermediate-view rendering used for scene-motion estimation; corresponding failure cases are discussed in Sec.~\ref{sec:limitations}. \textbf{DispFlow-GS consistently improves motion-aware performance on the evaluated scenes}, achieving the highest MMAP on both scenes and the best average MMAP and DRC. Based on the mean results, MMAP improves by approximately \textbf{4.5\%} over MotionGS and \textbf{12.5\%} over photometric-only Deformable 3DGS, while image-based performance remains comparable. Qualitatively, our method also \textbf{produces deformation that is more concentrated around the moving object} with fewer spurious responses in static regions, indicating that the proposed supervision generalizes beyond NeRF-DS while preserving reconstruction quality.

\subsection{Ablation Study}

We conduct ablations on NeRF-DS dataset to evaluate key components and design choices of our framework. Results are averaged across scenes and reported in Tabs.~\ref{tab:ablation_components}--\ref{tab:ablation_norm}.

\noindent \underline{\textbf{Key components.}}
Tab.~\ref{tab:ablation_components} evaluates Displacement Flow (DF), Motion Decomposition (MD), and Training Scheduling (TS). DF alone does not improve over the photometric baseline, indicating that changing the motion representation is insufficient when the supervision target still contains ambiguous motion. Once MD is introduced, MMAP increases by \textbf{47.7\%} and DRC by \textbf{12.5\%} over DF alone, showing that \textbf{motion decomposition is the key component that enables Displacement Flow to learn meaningful scene dynamics}. Without TS, this stronger motion constraint slightly degrades reconstruction quality; introducing TS restores the image-based metrics to approximately the baseline level while retaining most of the motion improvement, resulting in the final balance between motion and appearance.


\begin{table}[t!]
\centering
\caption{Ablation on key components: displacement flow (DF), motion decomposition (MD), and training scheduling (TS).}
  \vspace{-8pt}
\label{tab:ablation_components}
\resizebox{\linewidth}{!}{
\begin{tabular}{l|ccc|ccc|cc}
    \toprule
    Method & DF & MD & TS & PSNR$\uparrow$ & SSIM$\uparrow$ & LPIPS$\downarrow$ & MMAP$\uparrow$ & DRC$\uparrow$ \\
    \midrule
    Baseline & & & & 23.76 & 0.8485 & \best{0.1846} & 0.6497 & 0.6660 \\
    1 & \checkmark & & & 23.63 & 0.8406 & 0.1941 & 0.6306 & 0.6455 \\
    2 & \checkmark & & \checkmark & \second{23.77} & 0.8462 & 0.1878 & 0.6817 & 0.6577 \\ 
    3 & \checkmark & \checkmark & & 23.62 & \second{0.8467} & 0.1902 & \best{0.9316} & \best{0.7261} \\
    Ours & \checkmark & \checkmark & \checkmark & \best{23.80} & \best{0.8485} & \second{0.1869} & \second{0.9015} & \second{0.7081} \\
    \bottomrule
\end{tabular}}
\end{table}

\begin{table}[t!]
\centering
\caption{Ablation on motion representations and supervision targets.}
 \vspace{-8pt}
\label{tab:ablation_motion}
\resizebox{\linewidth}{!}{
\begin{tabular}{l|ccc|cc}
    \toprule
    Method & PSNR$\uparrow$ & SSIM$\uparrow$ & LPIPS$\downarrow$ & MMAP$\uparrow$ & DRC$\uparrow$ \\
    \midrule
    Gaussian Flow + Indirect Scene Motion (MotionGS) & 23.75 & 0.8290 & 0.2364 & 0.6825 & 0.6685 \\
    Gaussian Flow + Direct Scene Motion & 23.85 & 0.8485 & 0.1846 & 0.6317 & 0.6673 \\ 
    Displacement Flow + Indirect Scene Motion & 23.92 & 0.8498 & 0.1833 & \second{0.7258} & \second{0.6728} \\
    Displacement Flow + Direct Scene Motion (Ours) & 23.80 & 0.8485 & 0.1869 & \best{0.9015} & \best{0.7081} \\
    \bottomrule
\end{tabular}}
\vspace{-15pt}
\end{table}

\noindent \underline{\textbf{Motion representation and supervision target.}}
Tab.~\ref{tab:ablation_motion} further separates the effects of the motion representation and scene-motion target. Under the same indirect scene-motion supervision, replacing Gaussian Flow with Displacement Flow already improves MMAP, suggesting that it is \textbf{better aligned with deformation learning.} More importantly, switching from indirect to direct scene-motion estimation increases MMAP by \textbf{24.2\%} for Displacement Flow,  whereas it reduces MMAP for Gaussian Flow. This result shows that \textbf{Displacement Flow and direct scene-motion supervision are complementary}, with their combination giving the strongest motion-aware performance.


\noindent \underline{\textbf{Motion supervision settings.}}
Tab.~\ref{tab:ablation_settings} examines the flow backbone, motion-loss weight, decomposition order, and motion-mask filtering. RAFT~\cite{teed2020raft} gives nearly identical results to the default GMFlow~\cite{xu2022gmflow}, while MDFlow~\cite{kong2022mdflow} performs worse, suggesting that flow-prior quality matters but \textbf{the gains are not specific to GMFlow}; we retain GMFlow for fair comparison with MotionGS. Increasing the motion-loss weight improves motion metrics but sacrifices image quality, whereas reducing it weakens motion learning, supporting $\alpha=0.1$ as a balanced setting. Most notably, reversing the decomposition order reduces MMAP by approximately \textbf{38\%}, despite maintaining competitive image-based metrics. \textbf{The decomposition order is therefore critical}, as estimating scene motion under the photometrically constrained input viewpoint provides substantially more reliable supervision. Motion-mask filtering provides no consistent improvement, so it is omitted from the final model.


\begin{table}[t!]
\centering
\caption{Ablation on flow backbones, loss weights, and motion masks.}
 \vspace{-8pt}
\label{tab:ablation_settings}
\resizebox{\linewidth}{!}{
\begin{tabular}{l|ccc|cc}
    \toprule
    Method & PSNR$\uparrow$ & SSIM$\uparrow$ & LPIPS$\downarrow$ & MMAP$\uparrow$ & DRC$\uparrow$ \\
    \midrule
    Different flow model (RAFT \cite{teed2020raft}) & 23.78 & 0.8488 & 0.1847 & 0.9024 & 0.7083 \\
    Different flow model (MDFlow \cite{kong2022mdflow}) & 23.73 & 0.8476 & 0.1865 & 0.8084 & 0.6824 \\
    Smaller motion loss weight ($\alpha$ = 0.02) & 23.78 & 0.8478 & 0.1873 & 0.8206 & 0.6866 \\ 
    Larger motion loss weight ($\alpha$ = 0.5) & 23.61 & 0.8464 & 0.1910 & 0.9269 & 0.7216 \\
    Motion decomposition in reverse order & 23.87 & 0.8472 & 0.1892 & 0.5566 & 0.6473 \\
    Motion mask filtering & 23.78 & 0.8475 & 0.1889 & 0.8629 & 0.7063 \\
    Ours (GMFlow \cite{xu2022gmflow}; $\alpha$ = 0.1; w/o mask) & 23.80 & 0.8485 & 0.1869 & 0.9015 & 0.7081 \\
    \bottomrule
\end{tabular}}
\end{table}

\noindent \underline{\textbf{Normalization strategy.}}
Tab.~\ref{tab:ablation_norm} shows that \textbf{our method is relatively insensitive to the normalization strategy.} Interquartile-range normalization achieves the highest MMAP and DRC, but only modestly outperforms min-max, which gives slightly better image-based performance. We therefore adopt min-max for its simplicity and balanced performance across image- and motion-based metrics.


\begin{table}[t!]
\centering
\caption{Ablation study on motion normalization strategies.}
 \vspace{-8pt}
\label{tab:ablation_norm}
\resizebox{\linewidth}{!}{
\begin{tabular}{l|ccc|cc}
    \toprule
    Method & PSNR$\uparrow$ & SSIM$\uparrow$ & LPIPS$\downarrow$ & MMAP$\uparrow$ & DRC$\uparrow$ \\
    \midrule
    Max-Absolute Normalization & 23.75 & 0.8485 & 0.1869 & 0.8970 & 0.7041 \\
    Z-score Normalization & 23.39 & 0.8388 & 0.2024 & 0.8926 & 0.6973 \\
    Interquartile Range Normalization & 23.76 & 0.8483 & 0.1862 & 0.9254 & 0.7115 \\
    Min-Max Normalization (Ours) & 23.80 & 0.8485 & 0.1869 & 0.9015 & 0.7081 \\
    \bottomrule
\end{tabular}}
\vspace{-15pt}
\end{table}

%% file: sections/failure_cases.tex
\section{Discussion and Limitations}
\label{sec:limitations}

\noindent \textbf{Potential for Robotics.}
Robotic perception and spatial reasoning in dynamic environments require separating ego-motion from scene motion for reliable scene understanding. Our framework addresses this ambiguity by isolating scene motion under a fixed viewpoint and learning deformation concentrated on dynamic regions. Such motion-aware representations could support dynamic SLAM, navigation, and manipulation, where distinguishing persistent structure from transient motion is important. Our results also show that motion fidelity can improve substantially while image-based metrics remain nearly unchanged, suggesting that rendering quality alone may overlook properties relevant to robotic perception and spatial reasoning. Together, these results highlight the potential of motion-aware deformable 3DGS for robotic dynamic scene understanding and spatial reasoning.

\begin{figure}[t!]
\centering
\includegraphics[width=0.9\linewidth]{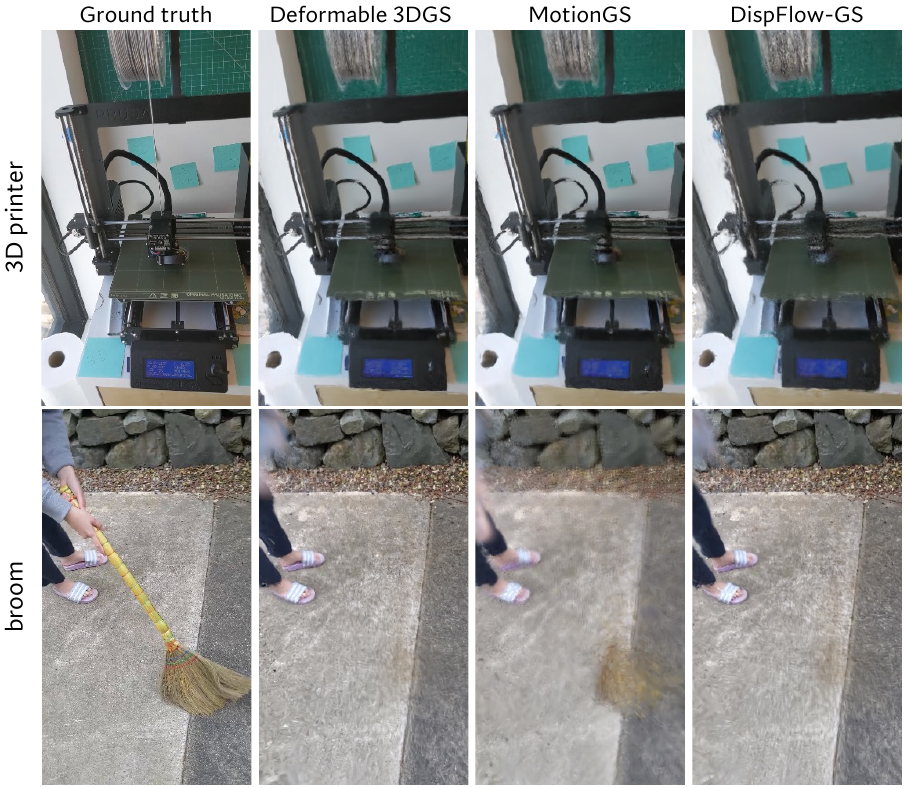}
\vspace{-8pt}
\caption{Failure cases on the HyperNeRF dataset.}
\label{fig:hypernerf_fail}
\vspace{-15pt}
\end{figure}

\noindent \textbf{Failure Cases.}
Failure cases occur in the \textit{``3D printer''} and \textit{``broom''} scenes of HyperNeRF (see Fig.~\ref{fig:hypernerf_fail}), mainly due to inaccurate camera poses, as also reported in Deformable\mbox{-}3DGS~\cite{yang2024deformable}. This leads to poor reconstruction of dynamic regions, such as the filament in \textit{``3D printer''} and the broom in \textit{``broom''}. Since our Displacement Flow supervision estimates scene motion from intermediate-frame rendering, it is particularly sensitive to pose errors.

\smallskip
\noindent \textbf{Limitations.}
Our framework is sensitive to optical-flow and camera-pose errors, affecting deformation learning and intermediate-frame rendering. Moreover, supervising motion magnitude rather than exact pixel correspondence may limit precise trajectory recovery in complex dynamic scenes.

%% file: sections/conclusion.tex
\section{Conclusion}
In this paper, we revisited motion supervision in Deformable 3DGS and showed that Gaussian flow is fundamentally limited by a domain mismatch with optical flow, weakening its ability to learn meaningful scene dynamics. We introduced Displacement Flow and scene--camera motion disentangling to provide more direct and domain-consistent supervision for deformation learning. Our experiments reveal a central finding: improved motion awareness does not necessarily translate into better photometric reconstruction or higher image-based metric scores. To better evaluate this discrepancy, we introduced Deformation--Rendering Consistency (DRC), which measures the alignment between predicted motion and rendering improvement. Together, these findings suggest that motion loss may offer limited benefit when image quality is the primary goal, whereas motion-aware deformation should be supervised and evaluated with motion-specific objectives and metrics. Building on these findings, future work will investigate task-oriented extensions of our motion disentangling and Displacement Flow for robotic dynamic scene understanding, with explicit integration into mapping and spatial-reasoning frameworks.